\pdfoutput=1

\documentclass[11pt]{article}

 \usepackage[final]{acl}
\usepackage{times}
\usepackage{latexsym}
\usepackage{colortbl}   

\usepackage[T1]{fontenc}

\usepackage[utf8]{inputenc}

\usepackage{microtype}

\usepackage{inconsolata}

\usepackage{graphicx}

\usepackage{algorithm}
\usepackage{algorithmic}
\usepackage{tikz}
\usepackage[most]{tcolorbox}
\usepackage{multirow}
\usepackage{adjustbox}
\usepackage{booktabs}
\usepackage{arydshln}
\usepackage{amsthm, mathtools,amsfonts}
\usepackage{amsmath}
\usepackage{amssymb}
\usepackage{soul}
\usepackage{graphicx}
\usepackage{todonotes} 
\usepackage{cleveref}
\usepackage{placeins}
\usepackage{threeparttable} 

\usepackage{xspace}

\newcommand{\GroundTruth}{\ensuremath{G}\xspace}   
\newcommand{\Generative}{\ensuremath{R}\xspace}    

\title{Do LLMs Adhere to Label Definitions? Examining Their Receptivity to External Label Definitions}

\author{
 \textbf{Seyedali Mohammadi\textsuperscript{1}},
 \textbf{Bhaskara Hanuma Vedula\textsuperscript{2}},
 \textbf{Hemank Lamba\textsuperscript{3}},
 \textbf{Edward Raff\textsuperscript{1,4}},
\\
 \textbf{Ponnurangam Kumaraguru\textsuperscript{2}},
 \textbf{Francis Ferraro\textsuperscript{1}},
 \textbf{Manas Gaur\textsuperscript{1}}
\\
\\
 \textsuperscript{1}UMBC,
 \textsuperscript{2}IIIT Hyderabad,
 \textsuperscript{3}Dataminr, Inc.,
 \textsuperscript{4}CrowdStrike
\\
 \small{
   {\{m294,edraff1,ferraro,manas\}@umbc.edu},vedula.hanuma@research.iiit.ac.in,
pk.guru@iiit.ac.in}
\\ \small{hlamba@dataminr.com}
}
\begin{document}
\maketitle
\begin{abstract} 
Do LLMs genuinely incorporate external definitions, or do they primarily rely on their parametric knowledge? To address these questions, we conduct controlled experiments across multiple explanation benchmark datasets (general and domain-specific) and label definition conditions, including expert-curated, LLM-generated, perturbed, and swapped definitions. Our results reveal that while explicit label definitions can enhance accuracy and explainability, their integration into an LLM’s task-solving processes is neither guaranteed nor consistent, suggesting reliance on internalized representations in many cases. Models often default to their internal representations, particularly in general tasks, whereas domain-specific tasks benefit more from explicit definitions. These findings underscore the need for a deeper understanding of how LLMs process external knowledge alongside their pre-existing capabilities.

\end{abstract}
\section{Introduction} 
\label{introduction}

\begin{figure}[ht]
  \centering
  \includegraphics[width=0.47\textwidth]{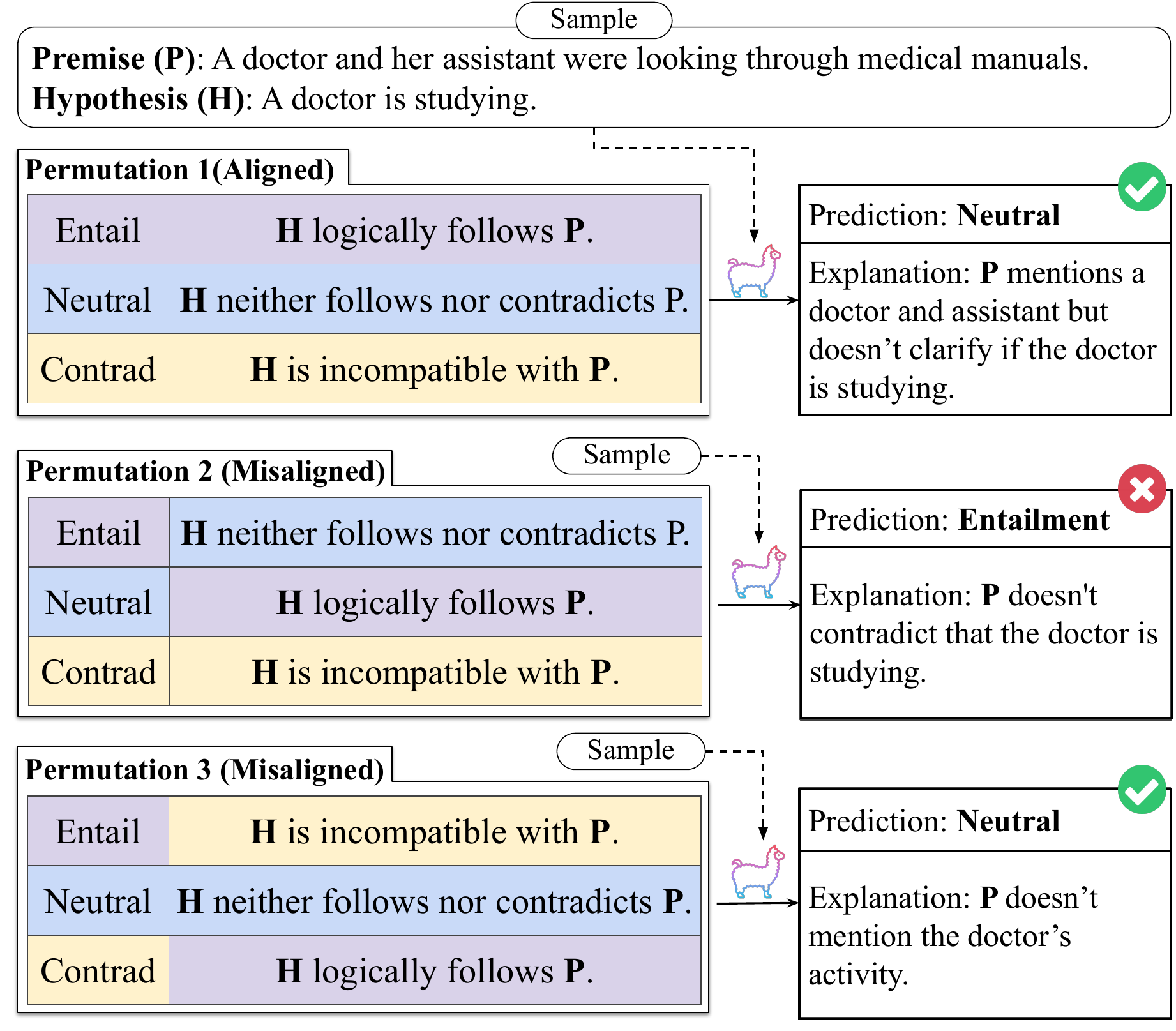}\caption{\footnotesize\label{fig:PermutationsFigure} \small The receptivity of LLMs to external definitions is revealed through a permutation analysis of the e-\textsc{SNLI} dataset. By testing all six possible orderings of entailment, neutral, and contradiction definitions, we discovered the model's receptivity varies significantly. In the illustrated example, the ground-truth label is ``neutral.'' Interestingly, in each permutation, the model consistently predicts the label that is mapped to the neutral definition, regardless of its original label name.}
\end{figure}

Label Definitions are considered as grounded statements that provide context on \textit{what an AI model} needs to do upon receiving a query ~\citep{mu-etal-2024-navigating,deng2025large}. These definitions are considered as clues to disambiguate unclear labels, helping models perform their tasks more effectively ~\citep{peskine2023definitions,kumar2023gen,xie2023adaptive}. However, whether models truly process and incorporate these definitions into their decision-making remains unclear. 

To highlight this, consider the example in \autoref{fig:PermutationsFigure}, where we test \textit{LLaMA-3}’s behavior under different permutations of label definitions on an instance of the e-SNLI dataset. When permutations 1 and 3 maintain the standard definition of \textit{``Neutral''} (cases where the premise neither entails nor contradicts the hypothesis) the model successfully classifies the relationship. However, when the \textit{``Neutral''} label is assigned an incorrect definition (permutation 2), \textit{LLaMA-3} misclassifies the input.  


This example reveals a fundamental challenge in how LLMs process conflicting information sources. When external definitions clash with a model's training data, we see unpredictable behavior; sometimes the model follows the external guidance, other times it relies on internal knowledge~\citep{burns2022discovering, duan2024llms}. To understand this systematically, we need to examine two key aspects: how models handle \textit{Knowledge Conflict} (when external definitions contradict training) and \textit{Definition Integration} (how definitions are presented in prompts)~\citep{azaria2023internal,gaur2022knowledge}. The \textit{LLaMA-3} case demonstrates both issues---the model's internal understanding of ``neutral'' conflicts with the manipulated definition, yet it still attempts to integrate the external guidance into its reasoning. This suggests that explanation-based classification may be more vulnerable to definitional inconsistencies than previously assumed.

To better understand the role of definitions in \textit{explanation-based classification} tasks using LLMs, we investigate the following questions: 
\textbf{\textit{Q1:}} Do LLMs rely more on their internalized, potentially opaque representations, raising questions about the interplay between external guidance and inherent knowledge? 
\textit{\textbf{Q2:}} Do they consistently adhere to external definitions? In this work, we restrict our study to classification tasks, as they provide clear ground truth labels that make it easier to assess whether models follow external definitions or revert to internal knowledge. We chose this focus deliberately: classification enables objective evaluation on widely used datasets (e.g., e-\textsc{snli}), establishes strong baselines for comparability, and reflects many real-world applications such as content moderation, diagnosis, or categorization. 
At the same time, our analysis of generated explanations already shows similar domain- and definition-sensitivity patterns, suggesting that the core findings may extend beyond classification to generation and reasoning tasks, which we leave for future work.

To answer these questions systematically, we conduct experiments across four diverse datasets: e-\textsc{snli} for general natural language inference~\citep{NIPS2018_8163}, \textsc{wellxplain} for mental health~\citep{garg2024wellxplain}, \textsc{hatexplain} for hate speech detection~\citep{mathew2020hatexplain}, and \textsc{wice} for fact-checking~\citep{kamoi2023wice}. This multi-domain approach allows us to test whether our findings generalize beyond any single task or domain. Our main contributions are as follows: 1) We develop evaluation strategies to assess the performance and explainability of LLMs when provided with label definitions, which serve as a lightweight and interpretable form of external knowledge injection. 2) We design two probing strategies, definition permutation and definitional perturbation, to assess the sensitivity of LLMs to definition quality and alignment. 3) We conduct extensive experiments across four diverse benchmarks, e-\textsc{snli}, \textsc{wellxplain}, \textsc{hatexplain}, and \textsc{wice}, to assess model receptivity to label definitions.

Our analysis yields four key insights: First, we show that models heavily rely on provided label definitions, with performance dropping significantly when definitions are swapped or corrupted. Second, we find that domain-specific tasks require more precise definitions than general tasks like natural language inference. Third, we demonstrate that LLM-generated definitions often outperform expert-written ones, particularly when they are tailored to specific inputs through simple retrieval methods, such as K-Nearest Neighbors (K-NN) \cite{sheth2021knowledge}. Finally, we reveal systematic differences in how models integrate definitions depending on the task domain and definition source. Code for reproducing our experiments is available at \url{https://github.com/mohammadi-ali/Definition-Receptivity-LLMs}.

\section{Problem Formulation}
\label{Sec:ProblemFormulation}

We evaluate the proclivity of an LLM $\mathcal{M}$ to adhere to a grounded definition\footnote{Grounded refers to definitions established by experts.} \GroundTruth and generative definition\footnote{Generative are those produced by LLM, e.g., \textit{GPT-4}.} \Generative through \textbf{Knowledge Conflict}, i.e., conflicting knowledge provided to the model and its internal knowledge, and through  \textbf{Definition Integration}, i.e., how to provide definitions to the model.

\subsection{Knowledge Conflict}
\label{Subsection:knowledgeConflict}
Conflicts arise when a model’s pre-trained knowledge conflicts with newly provided definitions, potentially affecting classification performance and explanation quality \citep{xie2023adaptive}. To assess the model's reliance on external definitions versus its internalized knowledge, we introduce two scenarios. 

Our first scenario examines \textit{varying degrees of definition accuracy} using three categories: \textit{incorrect}, \textit{slightly incorrect}, and \textit{correct} definitions. \textit{Incorrect} definitions completely misalign labels with contradictory explanations—for instance, defining ``contradiction'' in e-\textsc{snli} as a scenario where the hypothesis must always be true if the premise is true. \textit{Slightly incorrect} definitions introduce subtle inaccuracies that mislead without being entirely wrong—defining ``entailment'' as a relationship where the hypothesis is \textit{possibly} true given the premise, rather than \textit{necessarily} true. \textit{Correct} definitions use established logical relationships that align precisely with the intended classification task.

\noindent \textbf{Note.} We distinguish these categories from \emph{disputed definitions}, such as differing expert interpretations of terms like ``fluency'' or ``readability.'' These are not incorrect but represent legitimate divergences in perspective. While crucial in human evaluation settings, they introduce ambiguity that makes it difficult to isolate whether model performance changes arise from definition sensitivity or from inherent conceptual disagreement. For clarity, our study therefore focuses on clear right--wrong contrasts.

The second scenario, \textit{permutation-based}, involves swapping label definitions.
Given a set of $p$ label-definition pairs ($l_i, d_i$) as $\GroundTruth=\{(l_1,d_1),(l_2,d_2),\dots,(l_p,d_p)\}$. The set $\mathcal{P}$ contains all $p!$ possible permutations of $\GroundTruth$ which is denoted as the following:
\begin{equation*}
    \mathcal{P} = \{ \pi(\GroundTruth)|\pi\in S_p \}
\end{equation*}

where $S_p$ represents all possible ways to rearrange $p$ elements, containing $p!$ permutations. In this set, one unique permutation ($\pi_{c}$) represents the correct alignment, where each label matches its true definition, while the remaining $(p!-1)$ permutations ($\pi_{m}$) represent misaligned label-definition pairs. \\

\begin{figure*}[t]
  \centering
  \includegraphics[width=0.75\textwidth]{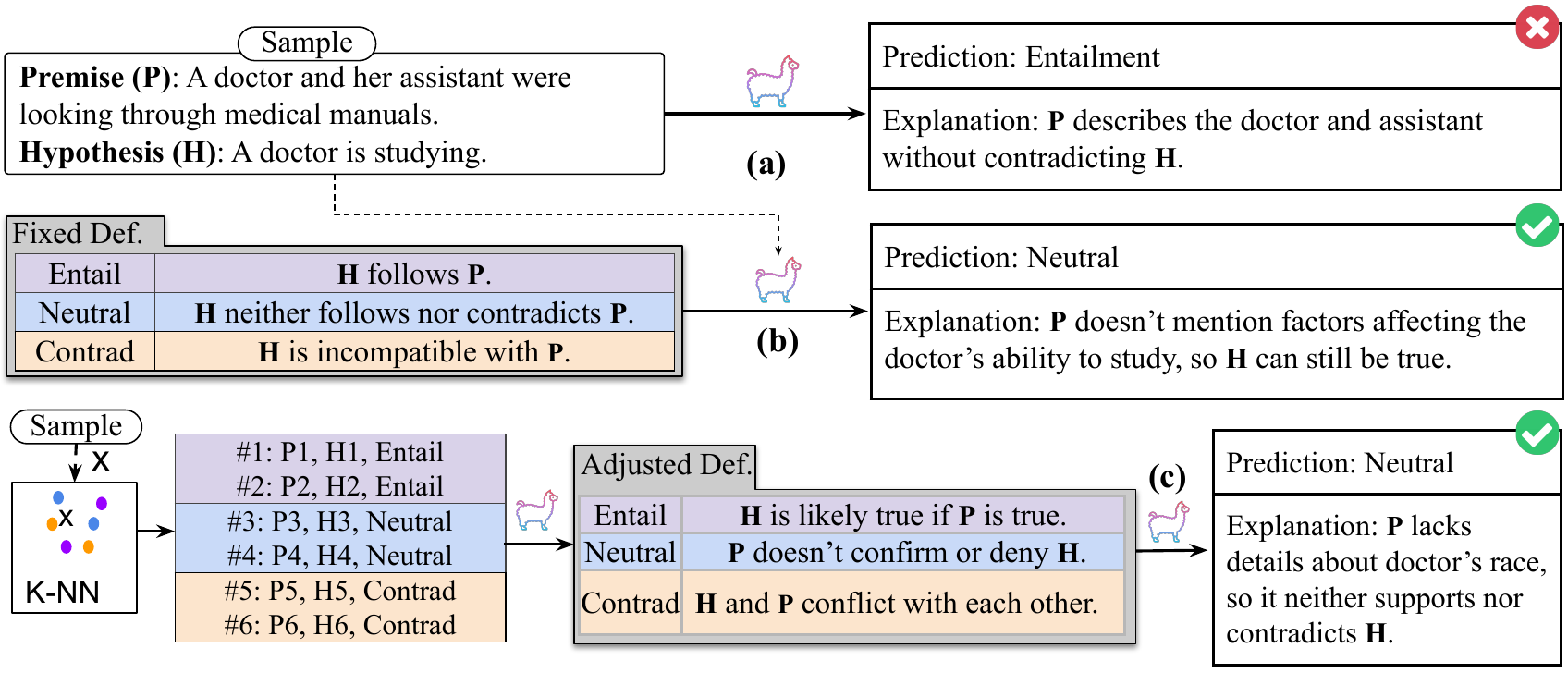}\caption{\footnotesize\label{fig:FixedAndAdjustedDefinitions} \small 
  An illustration of how \textit{LLaMA-3}'s natural language inference performance improves with label definition integration. (a) baseline performance where \textit{LLaMA-3} incorrectly labels a premise-hypothesis pair as ``Entailment''; (b) improved accuracy when grounded label definitions are provided, leading to correct ``Neutral'' classification; and (c) successful explanation and classification when presented with both definitions and few-shot examples. The same LLM (\textit{LLaMA-3}) is used to generate label definitions. The few-shot samples used to generate such definitions are in \autoref{tab:few-shot_samples} (\cref{additional_sec}). Note that the adjusted definitions shown in the figure are abbreviated; the complete versions are provided in \cref{adjust_label_definitons}.}
\end{figure*}

\noindent 
\textbf{Performance and Explanation Evaluation:} To assess how definition manipulation affects model performance, we use the Matthews Correlation Coefficient (MCC), a robust metric for binary classification that accounts for all four confusion matrix categories~\citep{10.1093/bioinformatics/16.5.412}. MCC ranges from -1 to +1, where +1 indicates perfect classification, 0 represents random performance, and -1 indicates complete disagreement between predictions and ground truth . Unlike accuracy or F1-score, MCC remains reliable even with imbalanced datasets.

For our permutation-based experiments, we compare performance when label-definition pairs maintain their correct associations, referred to as aligned mapping ($\text{MCC}_{a}$), against the average performance when these pairs are misaligned:

\vspace{-6pt}
\begin{equation*}
\overline{\text{MCC}}_m = \frac{1}{(p! - 1)} \sum_{\substack{i=1 \\ i \neq c}}^{p!} \text{MCC}_{\pi_i}
\end{equation*}

where $\pi_c$ represents the correct label-definition mapping and $\pi_i$ denotes the $i$-th permutation.

For definition accuracy experiments, we measure performance across three levels: $\text{MCC}_{\text{Inc.}}$ (incorrect definitions), $\text{MCC}_{\text{SInc.}}$ (slightly incorrect), and $\text{MCC}_{\text{Cor.}}$ (correct definitions). For most datasets (e-\textsc{snli}, \textsc{hatexplain}, \textsc{wellxplain}), models generate textual explanations that we evaluate using MCC. However, \textsc{wice} is a fact-checking dataset where explanations consist of supporting sentence indices rather than generated text. For this task, we report F1 scores calculated as:
\begin{equation*}
\text{F1} = \begin{cases}
1.0 & \text{if } |I_{\text{gold}}| = |I_{\text{pred}}| = 0 \\
0.0 & \text{if } |I_{\text{gold}}| = 0 \oplus |I_{\text{pred}}| = 0  \\
\frac{2 \cdot \text{P} \cdot \text{R}}{\text{P} + \text{R}} & \text{otherwise}
\end{cases}
\end{equation*}

Where $I_{\text{gold}}$ and $I_{\text{pred}}$ represent gold and predicted supporting sentence indices, and P and R are precision and recall, respectively.

\subsection{Definition Integration}

\label{Subsection:ContextualIntegration}

While our Knowledge Conflict experiments examine \textit{what happens} when definitions contradict model knowledge, Definition Integration investigates \textit{how} the presentation and source of definitions affects model behavior. 

We systematically evaluate four definition integration strategies using a test set $\mathcal{S}_{test} = \{(X_i, Y_i, E_i)\}_{i=1}^n$, where $X_i$ represents the input, $Y_i$ the true label, and $E_i$ the ground truth explanation. The four conditions are: \textit{(i) vanilla} (zero-shot), where models rely purely on internal knowledge without explicit definitions (\autoref{fig:FixedAndAdjustedDefinitions}(a)); \textit{(ii) fixed definition}, where expert-written definitions are incorporated directly into prompts (\autoref{fig:FixedAndAdjustedDefinitions}(b)); \textit{(iii) adjusted definition}, where definitions are dynamically generated for each input sample using the same LLM performing the classification task (\autoref{fig:FixedAndAdjustedDefinitions}(c)); and \textit{(iv) definition + few-shot}, which combines fixed definitions with exemplar cases. These conditions allow us to assess the trade-offs between relying on internal model knowledge versus external guidance, and between generic versus context-specific definitions. For most tasks, we evaluate explanation quality using ROUGE scores, while \textsc{wice} uses F1 scores for sentence-level explanation matching.

For the adjusted definition condition, we employ a k-NN approach to generate context-specific definitions that are tailored to each input. Given an input $X$, we retrieve the $k$ most similar training examples for each label $l$ using cosine similarity in embedding space, denoted as $\mathcal{N}_k(X, l)$. An LLM $\mathcal{M}$ then generates customized definitions based on these retrieved examples. The adjusted definitions are formalized as:

\begin{equation*}
    D_{adjusted} = \{\mathcal{M} (\mathcal{N}_k(X,l))\}_{l=1}^L
\end{equation*}

We experiment with $k \in \{1, 2, 5, 10\}$ and extend to 15 or 20 when necessary to identify optimal retrieval sizes. This approach allows us to test whether input-specific definitions, which potentially capture nuanced contextual information, improve both classification accuracy and explanation quality compared to static, expert-written definitions.

\section{Setup and Methodology}
\label{sec:methodology}
\paragraph{Models:} 
To investigate definition receptivity across diverse model architectures, we selected four state-of-the-art LLMs with deliberately varied characteristics: 
\textit{GPT-4},
\textit{LLaMA-3},
\textit{Phi-3}, and
\textit{Mistral-7B}.

This selection spans from compact 3.8B parameter models to trillion-parameter architectures, enabling us to determine whether investments in larger models yield proportional improvements in definition adherence—a key consideration for AI deployment in high-stakes domains where consistent interpretation of instructions is essential. Since definitions are concise, their inclusion adds minimal overhead in terms of token length. Nonetheless, this setup is primarily designed for short to medium-length inputs. For tasks requiring longer contexts (e.g., document-level inputs), the approach may not be suitable for smaller models like \textit{Phi-3}, which has a 4k token limit in our case, thereby restricting the ability to integrate external definitions without truncation.

Each model represents distinct design philosophies with practical implications for deploying LLMs in definition-sensitive contexts: \textit{GPT-4} serves as our high-performance benchmark, representing proprietary models that organizations might access through APIs~\citep{achiam2023gpt}; \textit{LLaMA-3} exemplifies open-source alternatives increasingly adopted in industry settings requiring customization and privacy~\citep{metaIntroducingMeta}; \textit{Phi-3} addresses the growing need for efficient, deployable models in resource-constrained environments~\citep{microsoftIntroducingPhi3}; and \textit{Mistral-7B} represents models optimized for the performance-efficiency trade-off crucial for commercial applications~\citep{jiang2023mistral}. This diversity allows us to provide actionable insights on which architectural approaches best ensure definitional consistency across different deployment scenarios—whether in cloud infrastructure, edge devices, or hybrid settings. Detailed specifications for each model are provided in \autoref{tab:implementation_details} (\cref{additional_sec}), with a representative prompt template shown in \autoref{tab:template_prompt} (\cref{additional_sec}). For \textit{GPT-4}, we balanced representation and cost by sampling 500 instances per label across datasets\footnote{This approach ensures balanced class representation while managing computational resources, a practical constraint in production settings as well.}. For all other models, we used complete test sets.

\begin{figure*}[t]
  \centering
  \includegraphics[width=0.85\textwidth]{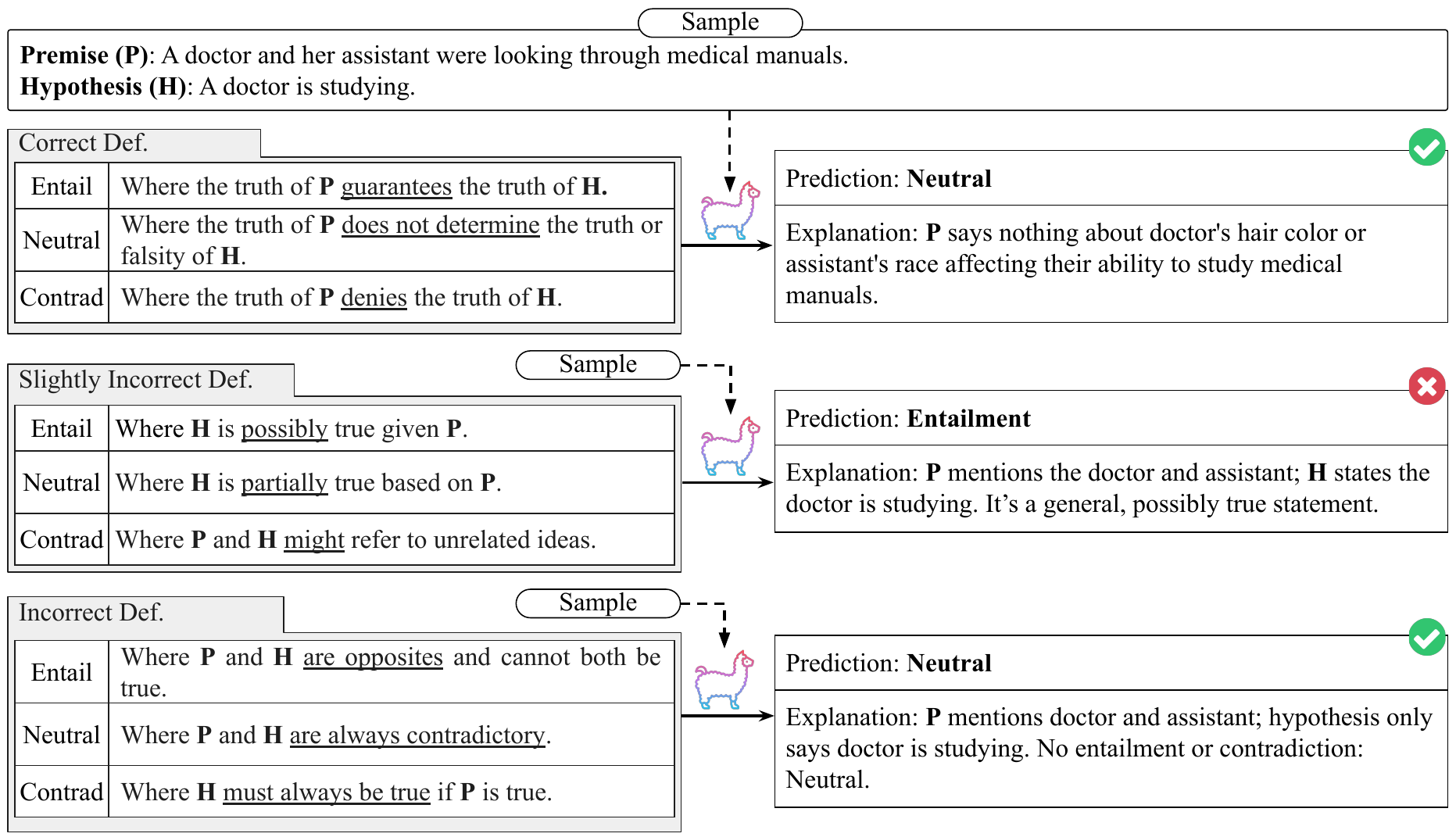}\caption{\footnotesize\label{fig:IncorrectVsSlightlyIncorrectVsCorrect} \small Label Definition Accuracy (Incorrect Vs. Slightly Incorrect Vs. Correct Definitions): As anticipated, when the model was provided with correct or slightly incorrect definitions, it produced the expected outputs: correct outputs for correct definitions and incorrect outputs for slightly incorrect definitions. However, when given incorrect definitions, the model generated correct outputs unexpectedly.}
\end{figure*}

\paragraph{Datasets:} 
Our experiments utilize four diverse datasets representing both general and domain-specific classification tasks. The e-\textsc{snli} dataset examines natural language inference with three labels: ``entailment'' (hypothesis necessarily follows from premise), ``neutral'' (hypothesis might be true given the premise), and ``contradiction'' (hypothesis cannot be true given the premise). The \textsc{wellxplain} dataset focuses on mental health categorization, classifying posts into four aspects: ``physical,'' ``intellectual and vocational,'' ``social,'' and ``spiritual and emotional'' well-being. \textsc{hatexplain} provides text classification across three categories: ``hatespeech,'' ``normal,'' or ``offensive'' content, representing typical content moderation scenarios. The \textsc{wice} dataset differs from the others as a fact-checking benchmark where models must identify which sentences from provided evidence support a given claim, requiring sentence-level explanation rather than just classification labels.

Statistical details of these datasets are presented in \autoref{tab:dataset} (\cref{additional_sec}), with the standard label definitions we use for fixed definition experiments provided in \cref{e-snli_labels,wellxplain-labels,hatexplain-labels,wice_definitions}. This diverse collection allows us to assess model behavior across varying tasks, from general logical reasoning to highly specialized domains where definition interpretation is particularly consequential.

\section{Experiments, Results, and Analysis}
\label{experiments_results}

\begin{table}[t]
    \centering
    \adjustbox{max width=\columnwidth}{
    \begin{tabular}{llllll}
    \toprule
    \rowcolor{gray!20}
    \textbf{Model}& \textbf{Permutation} &\textbf{e-\textsc{snli}}& \textbf{\textsc{wellxplain}}& \textbf{\textsc{hatexplain}} &\textbf{\textsc{wice}}\\
    \midrule
    \textit{GPT-4}& $\text{MCC}_a$ & \textbf{0.83}& \textbf{0.35} & \textbf{0.42 }&\textbf{0.32} \\
    & $\overline{\text{MCC}}_m$ & 0.74 &0.17 &0.27&0.16  \\
    \hline
     \textit{LLaMA-3}& $\text{MCC}_a$ & \textbf{0.42}&\textbf{0.20} &\textbf{0.21} &\textbf{0.13} \\
            & $\overline{\text{MCC}}_m $&0.34&0.15&0.17&0.09\\
     \hline
     \textit{Phi-3}& $\text{MCC}_a$ &\textbf{0.24}&\textbf{0.32}& \textbf{0.20} & \textbf{0.05}  \\
            &$\overline{\text{MCC}}_m$&0.23&0.11&0.06&0.04\\
     \hline
     \textit{Mistral-7B}& $\text{MCC}_a$ &\textbf{0.51}&\textbf{0.10}  &0.05&\textbf{0.05}\\
            &$\overline{\text{MCC}}_m$&0.48&0.07&0.05&0.03\\
     \bottomrule
     
    \end{tabular}
    }
    \caption{\footnotesize MCC for aligned vs. misaligned definitions across e-\textsc{snli}, \textsc{wellxplain}, and \textsc{hatexplain}. Aligned definitions improve performance, while misaligned ones reduce MCC. Subscripts \textit{a} and \textit{m} denote aligned and misaligned label–definition mappings, respectively.}
    \label{tab:permutation_results}
\end{table}

\begin{table*}[!ht]
    \centering
    \begin{threeparttable}
    \adjustbox{max width=\textwidth}{
    \begin{tabular}{lllllllllllll}
    \toprule
    \rowcolor{gray!20}
     &  \multicolumn{3}{c}{\textbf{e-\textsc{snli}}} & \multicolumn{3}{c}{\textbf{\textsc{wellxplain}}} &\multicolumn{3}{c}{\textbf{\textsc{hatexplain}}}&\multicolumn{3}{c}{\textbf{\textsc{wice}}}  \\ \cmidrule(lr){2-4} \cmidrule(lr){5-7} \cmidrule(lr){8-10} \cmidrule(lr){11-13}
    \rowcolor{gray!20}
   \textbf{Model} &  $MCC_{Inc.}$ & $MCC_{SInc.}$ & $MCC_{Cor.}$&  $MCC_{Inc.}$ & $MCC_{SInc.}$ & $MCC_{Cor.}$&  $MCC_{Inc.}$ & $MCC_{SInc.}$ & $MCC_{Cor.}$ &  $MCC_{Inc.}$ & $MCC_{SInc.}$ & $MCC_{Cor.}$ \\
    \midrule
       \textit{GPT-4}  & Meta-Response$^{\dagger}$   &0.83& 0.83 &  0.31& 0.31 & \textbf{0.38} &0.31 &0.36 &\textbf{0.44}&0.13&0.31&\textbf{0.32} \\ 
       \textit{LLaMA-3}  &  0.08 & 0.32 & \textbf{0.36} & 0.16 & 0.20 & \textbf{0.21} &0.10 &0.13 &\textbf{0.15}&0.09&0.13&0.13  \\
      \textit{Phi-3}    & 0.22 & 0.21 & \textbf{0.23}  & 0.29&0.35 &\textbf{0.38} & 0.11&0.08 &\textbf{0.14}&0.04&0.01&0.05  \\ 
      \textit{Mistral-7B}  &  0.20  & \textbf{0.48} & 0.40   &0.18 &\textbf{0.20} &\textbf{0.20} & 0.05 &0.07 &\textbf{0.09} &0.03&0.03&0.05 \\ \bottomrule
    \end{tabular}
    }
    \caption{\footnotesize Impact of Definition Quality: Incorrect, Slightly Incorrect, and Correct Definitions across the general domain dataset (e-\textsc{snli}) and domain-specific datasets (\textsc{wellxplain} and \textsc{hatexplain}). Results highlight distinct patterns in model performance, with \textit{LLaMA-3} showing steady improvement from Incorrect to Correct, while \textit{Mistral-7B} exhibits strong sensitivity to definition accuracy, particularly in domain-specific settings. Subscripts Inc., SInc., and Cor. denote Incorrect, Slightly Incorrect, and Correct, respectively. ``Meta-Response$^{\dagger}$'' reflects the model declining to answer due to conflicting definitions.\label{tab:incorrect_VS_conrrect_results}}
    \end{threeparttable}
    
\end{table*}

\subsection{LLM Receptivity: Results and Analysis}
\label{Subsection:Results}

Building on our experimental datasets, we now examine how different models integrate and respond to label definitions across our designed scenarios. Our analysis focuses on how models balance external definitions with their internalized knowledge.

In the \textit{definition permutation} scenario, \autoref{tab:permutation_results} shows clear performance differences between aligned and misaligned label-definition mappings. Across all datasets, models achieve significantly higher MCC scores with correctly permuted definitions (\textit{aligned}), with \textbf{\textit{Mistral-7B} on \textsc{hatexplain} being the striking exception} where alignment produced minimal improvement—a surprising deviation from the otherwise consistent pattern. These findings align with the distributional hypothesis of language learning~\citep{firth1957, saxena2024attribution}, whereby LLMs learn word meanings through their contextual associations. When faced with conflicting cues—explicit definitions versus internalized patterns—models generally prioritize the explicit definitions provided in the prompt context, suggesting that instruction-following capabilities can override pre-trained associations.

The \textit{label definition accuracy} scenario, summarized in \autoref{tab:incorrect_VS_conrrect_results}, reveals varying sensitivities to definition quality across models and tasks. For \textbf{e-\textsc{snli}}, \textit{LLaMA-3} demonstrates remarkably strong responsiveness, with MCC values increasing more than threefold when moving from \textit{incorrect} to \textit{slightly incorrect} definitions, and nearly fourfold when comparing \textit{incorrect} to \textit{correct} definitions. \textit{Mistral-7B} shows substantial but less dramatic improvements (140\% and 100\% respectively), while \textbf{\textit{Phi-3} exhibits a surprisingly minimal sensitivity} with near-negligible changes (-4.55\% and +4.55\%)—an unexpected resilience to definitional variations. Most unexpectedly, \textbf{\textit{GPT-4} surprised us entirely} by detecting definitional inconsistencies and declining to provide predictions—a sophisticated meta-cognitive response none of the other models exhibited. This pattern reveals a key distinction between linguistic competence (internalized knowledge of language structures) and performance (how that knowledge is applied in specific contexts), with larger models demonstrating stronger metacognitive capabilities to detect contradictions between provided definitions and their pre-training.

Notably, \textit{GPT-4} exhibits a unique \textit{meta-response} behavior, choosing abstention when faced with conflicting label definitions. This refusal, absent in other models, suggests that alignment strategies such as Reinforcement Learning from Human Feedback (RLHF) encourage diagnostic caution rather than committing to potentially inconsistent outputs. We emphasize this as an important extension of the competence–performance distinction and highlight it as a promising direction for mechanistic follow-up work on safety-aligned LLMs.

For the \textbf{\textsc{wellxplain}} mental health categorization task, all models show moderate improvements with better definitions, though the sensitivity varies. \textit{GPT-4} and \textit{LLaMA-3} show similar improvement patterns (22.58\% and 25\% respectively) when moving from incorrect to more accurate definitions. \textit{Phi-3} demonstrates greater definitional sensitivity, with performance improving by approximately one-fifth when using \textit{slightly incorrect} definitions and by nearly one-third with \textit{correct} definitions. \textbf{Contrary to expectations, \textit{Mistral-7B} shows the least variation} (11.11\%), suggesting it relies more heavily on internal knowledge for this domain—a surprising finding given its stronger responsiveness in the e-SNLI task. This aligns with theories of knowledge acquisition, suggesting that when LLMs lack extensive pre-training exposure to specialized domains, they become more dependent on explicit contextual definitions, similar to how humans learn novel concepts by relying on explicit instruction when they lack prior experience.

In the \textbf{\textsc{hatexplain}} content moderation task, definition quality impacts performance substantially across all models. \textit{GPT-4}'s accuracy improves steadily as definitions become more precise (16.13\% with \textit{slightly incorrect} and 41.94\% with \textit{correct} definitions). \textit{LLaMA-3} follows a similar pattern with more pronounced gains (30\% and 50\%). \textbf{Interestingly, both \textit{Phi-3} and \textit{Mistral-7B} show unexpectedly large improvements} when given correct definitions (75\% and 80\% respectively)—a stark contrast to their behavior on other tasks. This suggests a ``knowledge integration threshold'' where models shift from relying on internal representations to trusting external definitions when the latter are sufficiently precise and consistent with their training.

For the \textbf{\textsc{wice}} fact-checking task, model performance under varying definition accuracies reveals minimal responsiveness across all models. Unlike other datasets, the MCC scores remain nearly flat, with negligible gains when moving from incorrect to correct definitions. \textit{LLaMA-3} and \textit{Mistral-7B} show marginal improvement (from 0.09 to 0.13 and from 0.03 to 0.05, respectively), while \textit{Phi-3} even declines slightly from slightly incorrect to correct. These results suggest that \textbf{\textsc{wice}, as an evidence-retrieval task, is less sensitive to semantic label definitions} and more reliant on factual pattern matching—highlighting a limitation in current LLMs’ ability to incorporate abstract definitional guidance into decision-making for fact-based inference tasks.

These findings collectively reveal a dynamic interplay between pre-trained knowledge and contextual definitions that varies by model architecture, task domain, and definition quality. \textbf{Contrary to the conventional wisdom that larger models are universally more capable}, smaller models generally show higher definition-sensitivity, particularly in specialized domains, suggesting they have weaker internal representations and greater reliance on explicit guidance. \textbf{The most surprising finding is that model responses to definitions are not consistent across tasks}—the same model may be highly sensitive to definitions in one domain while showing minimal responsiveness in another. These results have important implications for prompt engineering in specialized applications, suggesting that careful definition crafting is crucial for smaller models and domain-specific tasks, while larger models benefit from their ability to incorporate definitions while maintaining critical evaluation of their validity. 

 To further understand how models incorporate external definitions into their decision-making, we next examine their receptivity across various definition integration strategies.
\begin{table*}[t]
    \centering
    \scriptsize
    \adjustbox{max width=\textwidth}{
    \begin{tabular}{lllllllllllllll}
    \toprule
    \rowcolor{gray!20}
     &  & \multicolumn{2}{c}{\textbf{e-\textsc{snli}}} & \multicolumn{2}{c}{\textbf{\textsc{wellxplain}}} & \multicolumn{2}{c}{\textbf{\textsc{hatexplain}}} &
     \multicolumn{2}{c}{\textbf{\textsc{wice}}}\\
     \cmidrule(lr){3-4} \cmidrule(lr){5-6} \cmidrule(lr){7-8} \cmidrule(lr){9-10}
    \rowcolor{gray!20}
    \textbf{Model}&\textbf{Scenario} & $MCC$ & $Rouge$ & $MCC$ & $Rouge$ &$MCC$ & $Rouge$ & $MCC$&$F1$ \\
    \midrule
                      & Vanilla    & \textbf{0.86}&0.21 & 0.41&  0.06& 0.33& 0.05&0.28&0.71\\
    \textbf{GPT-4}    & Fixed Definition  &0.80& 0.20 & 0.46& 0.05 & \textbf{0.36}&0.04&0.32&\textbf{0.73}\\ 
                      & Adjusted Definition  & 0.83& 0.20& \textbf{0.55}&0.07 & 0.33& 0.06&\textbf{0.35}&0.69\\
                      & Fixed Definition+Few-shot  & 0.70& \textbf{0.26}& 0.46&\textbf{0.10} & 0.30&\textbf{0.31}&0.32&0.73 \\

        \hline  
                      & Vanilla    &\textbf{0.47} & 0.18&0.29& 0.04 & 0.14&\textbf{0.31}&0.14&0.69\\
                     \textbf{LLaMA-3} & Fixed Definition  & 0.35&0.15 & 0.16&0.04 &\textbf{0.23}&0.18&0.13&0.69 \\ 
                   & Adjusted Definition  & 0.36& 0.14& \textbf{0.44}&\textbf{0.40} & 0.14&0.10&0.14&0.69\\
                    & Fixed Definition+Few-shot  & 0.36&\textbf{0.20}  & 0.38&0.38  & 0.15& 0.04&0.11&\textbf{0.74}\\
            
        \hline 
                      & Vanilla    & \textbf{0.48}&0.19 & 0.33&0.06 &0.13& 0.01&0.03&0.68\\
        \textbf{Phi-3} & Fixed Definition  &0.38&0.12  &0.38& 0.06 &\textbf{0.19}& 0.02&0.05&0.68\\ 
                      & Adjusted Definition  & 0.46& 0.15& 0.49&\textbf{0.14} & 0.17&0.01&\textbf{0.12}&0.67\\
                       & Fixed Definition+Few-shot  & 0.31&\textbf{0.22} &\textbf{0.62}&0.13  & 0.16&\textbf{0.10}&0.02&\textbf{0.69} \\

        \hline 
                      & Vanilla    &\textbf{0.57}& 0.17 &0.20&0.07  &0.01&\textbf{0.47} &0.03&0.70\\
        \textbf{Mistral-7B} & Fixed Definition  &0.32& 0.20 &0.20& 0.07 &0.12&0.35 &\textbf{0.05}&0.69\\ 
                         & Adjusted Definition  & 0.34&0.19 & \textbf{0.42}& \textbf{0.39}& \textbf{0.14}&0.01&0.04&0.69\\
                          & Fixed Definition+Few-shot  & 0.49& \textbf{0.24} & 0.39&0.36 & 0.14& 0.05 &0.01&0.70\\
    \bottomrule
    \end{tabular}
    }
    \caption{\footnotesize When external guidance is actually useful: definitions make general NLI harder, but they are very helpful in specialized domains, and adjusted definitions also improve performance in \textit{wice} for most models.
    }
    \label{tab:mcc_results}
\end{table*}

\subsection{Definitions Integration: Results and Analysis}
Building on our previous analysis of how models respond to definition conflicts, we now examine how different integration strategies affect performance. \autoref{tab:mcc_results} compares four conditions: (i) \textit{vanilla}, (ii) \textit{fixed definition}, (iii) \textit{adjusted definition}, and (iv) \textit{fixed definition + few-shot}.

For \textbf{e-\textsc{snli}}, \textbf{models perform best with the definition-free \textit{vanilla} setting—a counterintuitive finding} that challenges conventional wisdom about explicit guidance. While \textit{GPT-4} experiences modest degradation with definitions (6.98\% and 3.49\% decreases), \textit{LLaMA-3} and \textit{Mistral-7B} show more substantial performance drops (23.40\% to 43.86\%). This pattern aligns with the bayesian inference framework of in-context learning proposed by \citet{xieexplanation}, where models already possess robust internal representations of common linguistic tasks through pretraining and may experience interference when explicit definitions contradict these learned representations~\citep{min2022rethinking}. \textbf{Most surprisingly, adding few-shot examples alongside definitions further reduces performance across all models}, with \textit{GPT-4} dropping by 18.60\%—suggesting that exemplars can actually create confusion when combined with abstract definitions for general language tasks.

In contrast, for domain-specific tasks \textbf{\textsc{wellxplain}} and \textbf{\textsc{hatexplain}}, definitions generally enhance performance with some notable exceptions. \textit{GPT-4} benefits modestly (34.15\% in \textsc{wellxplain}, 9.09\% in \textsc{hatexplain}), while \textbf{Mistral demonstrates an extraordinary definition dependence with a 110\% improvement in \textsc{wellxplain} and a remarkable tenfold increase in \textsc{hatexplain}}. This dramatic improvement can be understood through the perspective of knowledge-base integration, where models with limited exposure to specialized domains during pretraining rely heavily on explicit definitions as surrogate knowledge, a phenomenon highlighted in prior work on definition-based supervision~\citep{wang2021entailment, mishra2022cross, mueller2022label}. \textit{LLaMA-3} exhibits task-dependent patterns, with \textbf{a peculiar reversal where \textit{fixed definitions} severely harm performance in \textsc{wellxplain}} (-44.83\%) despite helping considerably in \textsc{hatexplain} (+64.29\%). This inconsistency reflects what Reynolds and McDonell~\citep{reynolds2021prompt} describe as the brittleness of in-context learning, where slight variations in definition format can produce dramatically different outcomes depending on how well they align with a model's pretraining experiences. \textbf{Perhaps most unexpected, the smaller \textit{Phi-3} with \textit{fixed definition+few-shot} achieves an MCC of 0.62 for \textit{wellxplain}, outperforming all other models and conditions}—challenging assumptions about model scale advantages through what \citet{rafailov2023direct} termed the effectiveness of direct preference optimization in smaller models.

The explanation quality tells a different story from classification accuracy. \textit{Adjusted definition} dramatically improves Rouge scores in \textsc{wellxplain}, with \textbf{\textit{LLaMA-3} and \textit{Mistral-7B} showing unprecedented explanation improvements} (9$\times$ and 4.5$\times$, respectively). Similarly, \textit{fixed definition+few-shot} boosts explanation quality in \textsc{hatexplain} by \textbf{a staggering sixfold}. \textbf{Paradoxically, these dramatic improvements in explanation quality often fail to translate to higher classification accuracy}—revealing what \citet{kosinski2023theory} described as a distinction between a model's ability to reason about concept relationships (evident in explanations) versus its ability to apply those concepts in classification decisions, similar to the competence-performance distinction in human cognition~\citep{chomsky1965aspects}.

This dissociation highlights a deeper competence–performance gap in LLMs: while definitions enhance the models’ ability to articulate conceptual relationships in explanations (competence), they do not always shift the fast, categorical processes underlying prediction (performance). Our results suggest that explanation generation and classification rely on partially distinct subsystems, with external definitions influencing deliberative reasoning more readily than decision-making boundaries. Recognizing this gap is crucial for deploying LLMs in high-stakes applications where both accurate predictions and trustworthy explanations are required.

The fact-checking task (\textbf{\textsc{wice}}) reveals a different pattern. While \textit{GPT-4} shows modest benefits from definitions (25\% MCC improvement with \textit{adjusted definitions}), \textbf{all models demonstrate remarkably consistent F1 scores for explanation alignment regardless of definition condition}, a striking contrast to the variable impacts seen in other metrics and tasks. The stable explanation F1 on \textsc{wice} across definition conditions is consistent with its evidence-selection format (supporting sentence indices), which relies more on claim–evidence matching than label semantics\footnote{Minor wording/citation correction relative to the published proceedings version.}. 

These findings extend our understanding beyond the earlier knowledge conflict experiments by demonstrating that not only does definition sensitivity vary by model and domain, but \textbf{the specific integration strategy can dramatically affect both classification performance and explanation quality—sometimes in opposing directions}. The phenomenon mirrors what Wei et al.~\citep{wei2022emergent} termed ``emergent abilities'' in language models, where certain capabilities appear only under specific combinations of model architecture, task domain, and prompt structure—a finding with significant implications for how these models should be deployed in real-world applications.

These findings highlight the varying degrees to which LLMs integrate external definitions, with performance trends differing across general and domain-specific tasks.

\section{Related Work}
\label{related_works}

Label definitions have shown promise in improving zero-shot classification \cite{peskine2023definitions}. While the work establishes the basic utility of definitions, our study delves deeper into how LLMs process and utilize these definitions, particularly in generating explanations. Through our analysis of definition conditions, we reveal crucial insights into whether LLMs truly incorporate external definitions or default to their pre-trained knowledge.

While recent advances in reasoning methods, including chain-of-thought (CoT) prompting \cite{wang-etal-2024-analyzing, 10.1145/3639372, wei2023chainofthoughtpromptingelicitsreasoning}, Progressive-Hint Prompting \cite{zheng2024progressive}, and PromptWizard \cite{agarwal2024promptwizardtaskawarepromptoptimization}, have improved model performance, they cannot distinguish between genuine expert knowledge incorporation and superficial alignment \cite{greenblatt2024alignment}. Our work uniquely reveals how LLMs semantically process external definitions, offering insights into whether models truly incorporate or merely imitate expert knowledge.

Building on various prompt engineering techniques \cite{wu2024prompt, zhangautomatic, yang2023large, fernando2023promptbreederselfreferentialselfimprovementprompt} and explanation strategies \cite{lampinen-etal-2022-language}, our work provides the first systematic analysis of how LLMs interpret and integrate semantic definitions. This is particularly crucial for high-stakes applications like mental health and hate speech detection, where eliciting the model's true behavior rather than superficial compliance is essential—an aspect previous methods haven't fully explored.

While prior studies have examined knowledge conflicts in LLMs, our work is distinct in four ways. 
First, we introduce a dual framing that jointly considers both knowledge conflict and definition integration, whereas most work studies only conflict. 
Second, we provide the first systematic cross-domain comparison of definition receptivity, showing stark contrasts between general and domain-specific tasks. 
Third, we propose an exhaustive permutation-based probing strategy that tests all possible label-definition mappings, unlike prior binary conflict settings. 
Finally, we uncover an explanation–classification disconnect, revealing that explanation quality does not necessarily correlate with classification accuracy, a phenomenon not previously explored in this context.

In summary, our work is unique in four respects: 
(1) We jointly analyze both knowledge conflict and definition integration, whereas prior work has largely focused on conflict alone; 
(2) we provide the first cross-domain comparison, revealing that definition receptivity differs substantially between general and specialized domains; 
(3) we introduce a permutation-based probing strategy that exhaustively tests all possible label–definition mappings, beyond the binary conflict settings previously studied; and 
(4) we uncover a novel explanation–classification disconnect, showing that gains in explanation quality do not necessarily translate to improved classification accuracy. 
Together, these contributions differentiate our work from existing literature and open new directions for understanding LLM receptivity to definitions.

\section{Conclusion and Future Work}
\label{conclusion_future_work}
To evaluate LLMs' receptivity to label definitions, we study LLM across three domains—everyday language understanding, mental health analysis, and hate speech detection. We make the following findings:
Firstly, LLMs perform significantly better when handling definitions in specialized tasks than general everyday tasks. This indicates that their ability to follow instructions improves with domain-specific contexts. Based on this, we recommend crafting precise and context-specific definitions for specialized tasks such as mental health and hate speech. Clear definitions tailored to the task can greatly enhance model performance.

Secondly, the quality of definitions, whether correct or flawed, profoundly affects LLM behavior and output accuracy. Poorly constructed definitions can severely degrade model performance. Third, definitions customized for specific scenarios outperform generic, one-size-fits-all approaches. This highlights the importance of aligning definitions with the nuances of the task at hand. To maximize effectiveness, we recommend using a mix of fixed label definitions and a few shot examples that are semantically proximal to user queries in order  to adapt LLMs to the specific user context.

Finally, our observations hold true across various tasks ranging from general language understanding (e.g., entailment tasks) to specialized domains like mental health analysis and hate speech detection. This consistency underscores the versatility of LLMs but also reveals varying levels of task complexity. By adopting these strategies, domain experts and practitioners can unlock the full potential of LLMs across both general-purpose and specialized applications.

Our findings suggest actionable insights for deployment. On edge devices, smaller models (e.g., \textit{Phi-3}, \textit{Mistral-7B}) gain the most from carefully tailored definitions in specialized domains, offering lightweight performance boosts. In cloud settings, larger models (e.g., \textit{GPT-4}) rely more on internal knowledge but often refuse outputs when definitions conflict, making them better suited for high-stakes applications where safety is critical. Even when definitions do not improve classification, they reliably enhance explanation quality, supporting user trust, transparency, and regulatory compliance.

A promising future direction is to apply mechanistic interpretability techniques to understand better how external definitions influence model behavior. Such analysis can reveal whether specific components of the network are responsible for encoding and applying definition semantics, shedding light on the internal mechanisms that support or override external guidance during the tasks.

An additional avenue for future work is to examine LLM behavior under \emph{disputed but valid definitions}, where experts may legitimately disagree (e.g., on ``fluency'' vs. ``readability''). 

\section*{Limitations}
While our study provides strong evidence of the impact of label definitions on LLM performance, certain aspects warrant further exploration. Our analysis focuses on classification tasks, and while the results generalize well within this domain, additional studies in more complex reasoning tasks could provide deeper insights. Additionally, while our experiments rely on prompting strategies, future work could explore fine-tuning methods to further enhance definition adherence. Lastly, although we analyze definition accuracy and alignment, investigating finer-grained model behaviors through interpretability techniques could offer even more precise insights into how LLMs integrate external definitions. We did not employ attention-weight analysis, since prior work has shown that attention weights do not consistently reflect true model decision-making processes \citep{jain2019attention,wiegreffe2019attention,mohammadi-etal-2024-welldunn,naim2024explaining}. Nonetheless, we consider probing internal mechanisms, including attention, as valuable future directions.

\section*{Ethics Statement}
This research investigates the extent to which Large Language Models (LLMs) adhere to explicit label definitions. Our study is conducted using publicly available datasets and does not involve human subjects or personally identifiable information. While our analysis includes scenarios where models are presented with incorrect definitions, this is done solely to evaluate their robustness and receptivity to external knowledge. We acknowledge that LLMs may exhibit biases due to their pre-trained knowledge, and we do not make normative claims about their outputs. Our findings aim to contribute to the broader understanding of model alignment and interpretability for domain experts without promoting misuse or adversarial manipulation of AI systems.

\section*{Acknowledgments}
We thank Mohammad Eskandari for his valuable assistance in revising the figures presented in this paper. We also thank the anonymous reviewers for their constructive feedback and suggestions, which helped improve the quality of this work. We gratefully acknowledge support from the UMBC Cybersecurity Leadership – Exploratory Grant and the USISTEF Award. The opinions, conclusions, and recommendations expressed here are solely those of the authors and do not necessarily reflect the views of USISTEF, UMBC, or CrowdStrike.

This material is also based on research that is in part supported by the DARPA for the SciFy program under agreement number HR00112520301. The U.S. Government is authorized to reproduce and distribute reprints for Governmental purposes notwithstanding any copyright notation thereon. The views and conclusions contained herein are those of the authors and should not be interpreted as necessarily representing the official policies or endorsements, either express or implied, of DARPA or the U.S. Government.

\bibliography{acl_latex}

\appendix

\section{Additional tables}
\label{additional_sec}

\begin{table}[!ht]
\centering
\footnotesize
\adjustbox{max width=\columnwidth}{
\begin{tabular}{p{0.40\linewidth} p{0.40\linewidth} c}
\toprule
\rowcolor{gray!20}
\textbf{Premise} & \textbf{Hypothesis} & \textbf{Label} \\
\midrule
A doctor checks on his medical equipment. & A doctor preparing for work. & Entailment \\
\hline
A child is given a medical check up by a doctor. & A child with doctor. & Entailment \\
\hline
A doctor checks on his medical equipment. & A doctor getting ready for a patient. & Neutral \\
\hline
A child is given a medical check up by a doctor. & A sick child with doctor. & Neutral \\
\hline
A male doctor looking at a female patient’s hand. & The doctor is on a date. & Contradiction \\
\hline
A doctor in blue scrubs performing surgery. & A doctor is eating lunch. & Contradiction \\
\bottomrule
\end{tabular}}
\caption{Few-shot samples used to generate adjusted definitions (see \autoref{adjust_label_definitons} and \autoref{fig:FixedAndAdjustedDefinitions}c).}
\label{tab:few-shot_samples}
\end{table}


\begin{table}[!ht]
\centering
\adjustbox{max width=\columnwidth}{
\begin{tabular}{cccccc}
\toprule
\rowcolor{gray!20}
  \textbf{Model} & \textbf{Version} &\textbf{\# parameters}     \\
\midrule
 \textit{GPT-4} & gpt-4-0613 &-   \\
 \textit{LLaMA-3} & \href{https://huggingface.co/meta-LLaMA/Meta-LLaMA-3-8B-Instruct}{meta-LLaMA/Meta-LLaMA-3-8B-Instruct}  & 8B  \\
\textit{Phi-3} & \href{https://huggingface.co/microsoft/Phi-3-mini-4k-instruct}{microsoft/Phi-3-mini-4k-instruct} & 3.82B    \\
 \textit{Mistral-7B} & \href{https://huggingface.co/mistralai/Mistral-7B-Instruct-v0.1}{mistralai/Mistral-7B-Instruct-v0.1}& 7.24B    \\
\bottomrule
\end{tabular}
}
\caption{\label{tab:implementation_details} Models' detail used in experimental setup.}
\end{table}

\begin{table}[!ht]
\centering
\adjustbox{max width=\columnwidth}{
\begin{tabular}{p{\linewidth}}
\toprule
\rowcolor{gray!20}
\multicolumn{1}{c}{\textbf{Template Prompt}} \\ 
\midrule
\textbf{Prompt:} Read the following definitions for neutral (1), entailment (0), and contradiction (2) and come up with a label and explanation for the given promise-hypothesis pair:\\
\textit{Label Definitions}: \{definitions (fixed or adjusted) here\}\\
\textit{Premise}: \{premise here\}\\
\textit{Hypothesis}: \{hypothesis here\}\\

\hline
\textbf{Response:} \\
\textit{Predicted Label}: \{your label here\}\\
\textit{Explanation}: \{your explanation here\}\\
\bottomrule
\end{tabular}}
\caption{Template Prompt Used for \textit{Mistral-7B} in the Definition Scenario, for e-\textsc{snli}. For other scenario prompts, refer to the corresponding code.}
\label{tab:template_prompt}
\end{table}

\begin{table}[!ht]
\footnotesize
\centering
\adjustbox{max width=\columnwidth}{
\begin{tabular}{ccccccc}
\toprule
\rowcolor{gray!20}
  \textbf{Dataset} & \textbf{Total}&\textbf{Train} & \textbf{Valid}&\textbf{Test}  \\
\midrule
 e-\textsc{snli} &569,051&549,367 &9,824&9,842  \\
 \textsc{wellxplain} &3,092 &-&-&- \\
\textsc{hatexplain} &19,229&15,383&1,922&1,924\\
\textsc{wice}(claim) &1967&1260&349&358\\
\bottomrule
\end{tabular}
}
\caption{\label{tab:dataset} Statistical summary of e-\textsc{snli}, \textsc{wellxplain}, \textsc{hatexplain}, and \textsc{wice} datasets.
}
\end{table}
\section{Definitions of labels in e-\textsc{snli}}
\label{e-snli_labels}
\begin{itemize}
    \item \textbf{Entailment (0):} if the premise entails the hypothesis.
    \item \textbf{Neutral (1):} if neither entailment nor contradiction hold.
    \item \textbf{Contradiction (2):} if the hypothesis contradicts the premise.
    
\end{itemize}

\section{Definitions of labels in \textsc{wellxplain}}
\label{wellxplain-labels}
\begin{itemize}
    \item \textbf{Physical Aspect (PA)}: Physical wellness fosters healthy dietary practices while discouraging harmful behaviors like tobacco use, drug misuse, and excessive alcohol consumption. Achieving optimal physical wellness involves regular physical activity, sufficient sleep, vitality, enthusiasm, and beneficial eating habits. Body shaming can negatively affect physical well-being by increasing awareness of medical history and appearance issues.
    \item \textbf{Intellectual Aspect (IA)}: Utilizing intellectual and cultural activities, both inside and outside the classroom, and leveraging human and learning resources enhance the wellness of an individual by nurturing intellectual growth and stimulation. 
    \item \textbf{Vocational Aspect (VA)}: The Vocational Dimension acknowledges the role of personal gratification and enrichment derived from one's occupation in shaping life satisfaction. It influences an individual's perspective on creative problem-solving, professional development, and the management of financial obligations.
    \item \textbf{Social Aspect (SA)}: The Social Dimension highlights the interplay between society and the natural environment, increasing individuals' awareness of their role in society and their impact on ecosystems. Social bonds enhance interpersonal traits, enabling a better understanding and appreciation of cultural influences.
    \item \textbf{Spiritual Aspect (SpA)}: The Spiritual Dimension involves seeking the meaning and purpose of human life, appreciating its vastness and natural forces, and achieving harmony within oneself.
    \item \textbf{Emotional Aspect (EA)}: The Emotional Dimension enhances self-awareness and positivity, promoting better emotional control, realistic self-appraisal, independence, and effective stress management.
\end{itemize}

\section{Definitions of labels in \textsc{hatexplain}}
\label{hatexplain-labels}
\begin{itemize}
    \item \textbf{Hatespeech:} is used to describe posts that express hatred towards a specific group or individual based on attributes like race, religion, gender, or sexual orientation. It is defined in the context of online content that devalues or discriminates against minority members and can lead to increased prejudice and real-world violence against these groups.
    
    \item \textbf{Normal:} Posts labeled as normal are those that do not contain any hate speech or offensive content. These posts are considered benign and do not target any individual or group with harmful or abusive language.

    \item \textbf{Offensive:} The term ``offensive'' is used to describe language that is abusive or insulting but does not necessarily meet the criteria of hate speech. Offensive speech includes harmful or derogatory language that can hurt individuals or groups but lacks the broader discriminatory intent characteristic of hate speech. 
\end{itemize}

\section{Definitions of labels in \textsc{wice}}
\label{wice_definitions}
\begin{itemize}
    \item \textbf{Supported:} Entire claim is backed by evidence.
    \item \textbf{Partially supported:} Some parts are supported, others not.
    \item \textbf{Not supported:} No part is supported.
\end{itemize}

\section{Adjusted label definitions for sample in \autoref{fig:FixedAndAdjustedDefinitions}(c)}
\label{adjust_label_definitons}
\begin{itemize}
    \item \textbf{Entailment:} is the relationship between a premise and a hypothesis where the truth of the premise logically guarantees the truth of the hypothesis, often indicating a necessary condition or a step in a process.
    \item \textbf{Neutral:} refers to a statement or situation that lacks a clear positive or negative connotation, implication, or emotional tone, often indicating a lack of direct or explicit relationship between the premises and the hypothesis.
    \item \textbf{Contradiction:} A contradiction is a statement that cannot be true or valid because it involves two or more mutually exclusive or incompatible circumstances, actions, or facts.
\end{itemize}

\end{document}